\documentclass{article}

 \usepackage{arxiv}

 \usepackage[utf8]{inputenc} 
 \usepackage[T1]{fontenc}    
 \usepackage[numbers]{natbib}
 \usepackage{url}            
\usepackage{booktabs}       
\usepackage{amsfonts}       
\usepackage{nicefrac}       
\usepackage{microtype}      
\usepackage{xcolor}         
\usepackage{graphicx}       
\usepackage{enumitem}       
\usepackage{hyperref}       
\hypersetup{
  hidelinks,
  pdftitle={We Ran a Research Sprint With AI. The AI Worked. The Research Didn't.},
  pdfauthor={Anush Sankaran and Arijit Banerjee}
}

\title{We Ran a Research Sprint With AI. \\ The AI Worked. The Research Didn't.}
\renewcommand{\shorttitle}{We Ran a Research Sprint With AI}

\author{%
  Anush Sankaran \\
  \textnormal{Microsoft}
  \And
  Arijit Banerjee \\
  \textnormal{Microsoft}
}
\date{}

\begin{document}

\maketitle

\begin{abstract}
Generative AI tools for research are widely deployed, yet individual productivity gains do not guarantee better team research. Following \citet{farach2026scaffolding}, we argue that how AI use is structured matters alongside access. Personal AI tools may amplify two failures under deadline pressure: a \emph{consumption bottleneck}, in which generation outpaces team-visible evaluation, and \emph{AI-induced research silos}, in which private human-AI loops displace shared decision-making. We motivate this position with a mechanism-generating pilot: a 100-minute research sprint with twelve researchers in three independently operating teams, within a seventeen-person session. Both patterns recurred across teams and evidence sources. Of 33 logged AI interactions, one was accepted outright and 27 lacked a recorded disposition; missing records alone do not establish absent evaluation. A parallel autonomous pipeline produced a complete draft without in-session human evaluation, illustrating production capacity rather than comparative quality. We propose \emph{Human-AI-Human} collaboration: an interaction topology in which AI connects researchers' contexts and supports inspectable collective evaluation and decision-making. The pilot does not isolate AI-specific effects from time pressure or workflow practices. We close with seven modular, falsifiable design hypotheses, prioritizing shared context and team-visible evaluation checkpoints for controlled follow-up.
\end{abstract}

\section{Introduction}
\label{sec:introduction}

Studies document individual productivity gains alongside task-dependent limitations \citep{brynjolfsson2023genaiwork,dellacqua2023jagged,vaccaro2024metaanalysis}. \citet{farach2026scaffolding}, in a field experiment with 388 employees at a Fortune 500 retailer, argue that the structural conditions surrounding AI use matter alongside access. Prior work examines individual coding and writing assistance \citep{vaithilingam2022expectation,lee2022coauthor}, agentic execution \citep{wu2023autogen,boiko2023coscientist}, and automated machine learning drafts \citep{lu2024aiscientist,yamada2025aiscientistv2}. AI-generated research ideas have been rated more novel, but slightly less feasible, than expert ideas \citep{si2024novel}; controlled writing and coding studies report productivity gains \citep{noy2023productivity,peng2023copilot}. These results motivate testing, rather than assuming, improvements in team research.

We focus on what changes when the unit of work is a team. Human-AI interaction guidelines \citep{amershi2019guidelines} and empirical design-space surveys \citep{lai2023science} provide foundations for studying people working with AI. Our question concerns shared context and collective evaluation when researchers work in parallel, hand off context, and make consequential decisions together. The CSCW and team-cognition literatures offer relevant concepts: common ground \citep{clark1991grounding}, workspace awareness \citep{gutwin2002workspace}, distributed cognition \citep{hollan2000distributed}, transactive memory \citep{wegner1987transactive}, and observability and directability when automation joins a team \citep{klein2004ten,seeber2020machines}.
In parallel, AI access has been folded into expectations of speed without renegotiating verification costs. AI-generated phrasing has diffused measurably across scientific papers \citep{liang2024llmphrasing}, although this does not establish declining quality. We use \emph{research slop} for the broader risk of generation outpacing evaluation, not an outcome measured here. Table~\ref{tab:trichotomy} distinguishes three interaction topologies. Our pilot motivates investigating the third: AI supporting shared context and collective evaluation rather than only individual production.

\begin{table}[t]
\centering
\small
\caption{The three-mode design space the field is navigating.}
\label{tab:trichotomy}
\begin{tabular}{@{}p{2.0cm} p{3.0cm} p{2.4cm} p{3.8cm}@{}}
\toprule
\textbf{Mode} & \textbf{Unit of analysis} & \textbf{Status} & \textbf{Evaluation concern} \\
\midrule
Human-AI & One person, one model & Default & Private contexts may fragment team evaluation. \\
AI-AI & One pipeline, no human in loop & Trajectory \citep{lu2024aiscientist,yamada2025aiscientistv2} & Human verification deferred downstream. \\
Human-AI-Human & The team; AI mediates between researchers & Design agenda & Collective evaluation remains to be tested. \\
\bottomrule
\end{tabular}
\end{table}

\begin{figure}[t]
  \centering
  \includegraphics[width=1\linewidth]{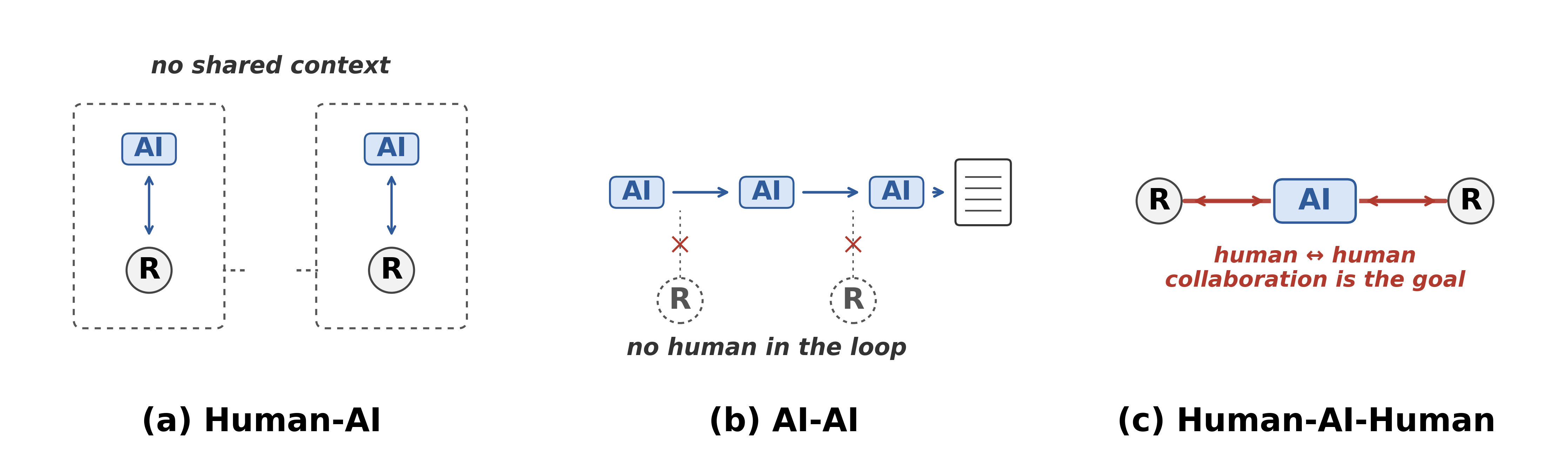}
  \caption{The three-mode design space described visually.}
  \label{fig:hah}
\end{figure}

We conducted a 100-minute instrumented pilot with twelve researchers in three independently operating teams, each paired with a silent observer, within a seventeen-person session. Teams were asked to produce an empirical paper with access to Claude, Claude Code, ChatGPT, and Copilot. Observers logged AI interactions and dispositions; stage-transition micro-surveys and a closing adversarial debrief captured participant reflections. A parallel autonomous pipeline produced a draft without human intervention, serving as a limit case rather than a quality comparison. Across teams, reflections, logs, and artifacts suggested two recurring patterns: a \emph{consumption bottleneck}, where generation outpaced collective evaluation, and \emph{AI-induced research silos}, where private sessions impeded shared understanding. One participant wrote: \textit{``AI is a multiplier for personal work, not team work.''} These convergent observations motivate hypotheses, not causal estimates.

\paragraph{Proposed Position:} \textbf{AI tools designed as personal multipliers may amplify consumption bottlenecks and private research silos under deadline pressure. We argue for \emph{Human-AI-Human} systems that connect researchers' contexts, expose disagreements, and support collective evaluation. Better workflows and existing collaboration platforms may realize this category; stronger models alone do not guarantee it. Collective evaluation is a testable pathway toward research quality, not an improvement demonstrated by this pilot.}

\paragraph{Contributions.} We characterize two candidate team-level mechanisms, a \emph{consumption bottleneck} and \emph{AI-induced silos} (\S\ref{sec:findings}); distinguish autonomous production from human evaluation (\S\ref{sec:findings:autonomous}); organize interaction topologies (Table~\ref{tab:trichotomy}); and propose seven falsifiable design hypotheses for team-centered AI (\S\ref{sec:gap}).
\section{Related Work}
\label{sec:related}

We map five clusters relevant to AI-assisted research, distinguishing evidence about individual assistance, autonomous execution, and team collaboration.

\paragraph{AI as research worker.}
Tool-using and agentic systems \citep{yao2023react,wu2023autogen,hong2024metagpt} have produced scientific instantiations including Coscientist \citep{boiko2023coscientist}, FunSearch \citep{romera2024funsearch}, end-to-end paper pipelines \citep{lu2024aiscientist,yamada2025aiscientistv2}, and Agent Laboratory \citep{schmidgall2025agentlab}. Galactica targets scientific knowledge generation \citep{taylor2022galactica}. These systems establish production capabilities, not necessarily shared team evaluation. Human participation is not categorically absent: AutoGen supports human inputs, and Agent Laboratory enables feedback at each research stage.

\paragraph{Generative AI and knowledge workers.}
Productivity gains are documented in several settings \citep{noy2023productivity,brynjolfsson2023genaiwork,peng2023copilot,cui2024devs}, but the qualifications are substantive. \citet{dellacqua2023jagged} report poorer performance outside the model's frontier. \citet{vaccaro2024metaanalysis} find that human+AI underperforms the better of either alone on average, with important task differences. In short-story writing, \citet{doshi2024creativity} find higher individual creativity but reduced diversity across outputs; this is an analogy, not evidence about research teams. Metacognitive demands \citep{tankelevitch2024metacognitive,subramonyam2024gulf} and suggestion-verification costs \citep{mozannar2024reading} motivate our consumption-bottleneck hypothesis. Whether shared research artifacts can sustain collective evaluation across private AI sessions remains a question.

\paragraph{Scaffolding AI use at the team level.}
The most directly relevant recent evidence is \citet{farach2026scaffolding}, a field experiment with 388 employees at Gap Inc.\ that varied how AI use was structured. Their synchronous Create-Out-Loud protocol (pairs discussed verbally, then prompted Copilot to draft from the transcript) was associated with documents scoring 4.96 points lower on a 22-point rubric ($p<.001$) and lower odds of producing a document (OR $=0.12$). Session timing, differential attrition, and length-sensitive grading complicate causal interpretation. Reframing AI as a thought partner showed gains concentrated in the upper quality distribution. At Procter \& Gamble, individuals with AI matched unaided teams on innovation tasks \citep{dellacqua2025cybernetic}. A medical field study links ``engaged augmentation'' to professionals actively interrogating AI outputs \citep{lebovitz2022engage}. These findings motivate testing how AI supports collective evaluation, not prescribing who speaks first. Human-AI-Human concerns shared context and decision authority, rather than a particular ordering of conversation and drafting.

\paragraph{Human-AI interaction and co-creation.}
Interaction guidelines \citep{amershi2019guidelines} and decision-making surveys \citep{lai2023science} provide design foundations. Appropriate reliance depends on explanation design, cognitive effort, and automation conditions \citep{bansal2021whole,vasconcelos2023explanations,bucinca2021forcing,parasuraman2010complacency,passi2022overreliance}. CoAuthor documents individual writer-model interaction \citep{lee2022coauthor}; a broader design space spans writing-assistant tasks and interactions \citep{lee2024writingdesignspace}; VISAR supports argument planning and draft prototyping \citep{zhang2023beyond}. Opinionated models can shift users' views \citep{jakesch2023opinionated}, motivating our hypothesis that private sessions could alter team alignment.

\paragraph{Team cognition, CSCW, AI in groups.}
This cluster motivates questions about common ground \citep{clark1991grounding}, workspace awareness \citep{gutwin2002workspace}, distributed cognition \citep{hollan2000distributed}, transactive memory \citep{wegner1987transactive}, observable and directable automation \citep{klein2004ten,seeber2020machines}, and AI-mediated communication \citep{hancock2020aimediated}. Automation can leave humans with demanding residual roles \citep{bainbridge1983ironies}. Group-AI work is suggestive: LLM devil's advocates can promote appropriate group reliance on AI \citep{chiang2024devils}; LLM-augmented brainwriting explores group ideation and evaluation \citep{shaer2024brainstorming}; selectively requesting peer second opinions can mitigate AI overreliance \citep{lu2024secondopinions}; and AI suggestions improved empathy in text-based peer support \citep{sharma2023empathic}. \citet{messeri2024illusions} discuss risks of illusions of understanding in science; our position examines candidate organizational mechanisms.

\paragraph{Nearest neighbors.}
Table~\ref{tab:neighbors} distinguishes seven related works from our proposed focus on shared context and team-visible evaluation.

\begin{table}[t]
  \centering
  \small
  \caption{Nearest neighbors and what we add.}
  \label{tab:neighbors}
  \begin{tabular}{@{}p{2.6cm} p{6.0cm} p{6.6cm}@{}}
    \toprule
    \textbf{Prior work} & \textbf{Claim} & \textbf{What we add} \\
    \midrule
    \citet{farach2026scaffolding} & A synchronous AI drafting protocol was associated with lower document quality and production. & A motivation to test shared context and collective evaluation, not a fixed sequence of human and AI turns. \\
    \citet{tankelevitch2024metacognitive} & A conceptual framework for individual metacognitive demands of generative AI. & A proposed team-level bottleneck when shared evaluation competes with production. \\
    \citet{messeri2024illusions} & AI may foster illusions of understanding in science. & Candidate team-level organizational mechanisms. \\
    \citet{vaccaro2024metaanalysis} & Human+AI is not automatically better than the best of either. & Shared context and collective evaluation as specific team-level design variables. \\
    \citet{liang2024usability} & Developers report difficulty controlling outputs and meeting code requirements. & Cross-person context and collective evaluation as additional design concerns. \\
    \citet{chiang2024devils} & LLM devil's advocates can promote appropriate group reliance on AI. & A broader topology encompassing this intervention. \\
    \citet{wu2023autogen} & Multi-agent conversations can combine models, tools, and human inputs. & Explicit support for shared researcher contexts and collective decisions. \\
    \bottomrule
  \end{tabular}
\end{table}
\section{The Session as Instrument}
\label{sec:method}

This mechanism-generating pilot involved seventeen people: twelve applied AI researchers in three four-person teams, three non-participating AI-researcher observers, an orchestrator, and a critic. Recruitment was voluntary from an internal organizational cohort; task participants were already familiar with the available AI tools. Participants provided written informed consent covering possible publication, and the study underwent internal organizational research review. The shared task was to produce an empirical paper on \emph{how do we capture the evolution of a research idea, and find the gaps in it, in a human-agent setup?}

The 100-minute session imposed hard cutoffs across ideation, design, execution, analysis, and writing. Teams chose among prompt styles (A~Direct, B~Collaborative, C~Roleplay, D~Adversarial) and freely used Claude, Claude Code, ChatGPT, or Copilot, without facilitation on prompting or division of labor. A silent observer per team logged interactions and dispositions (accepted / modified / rejected / unresolved). Reliance research informed this focus \citep{passi2022overreliance,vasconcelos2023explanations}; the four categories were our operationalization. Short stage-transition surveys and a closing adversarial debrief supplied complementary evidence. Participants described AI as useful for literature review, refining ideas, evaluation criteria, and executing specified plans, but difficult to consume, verify, and consolidate under deadline.

\paragraph{Why 100 minutes, and the autonomous parallel run.}
The compressed deadline made end-to-end observation feasible; it cannot establish whether pressure exposed or created the patterns. Teams chose B/C/D for ideation, describing adversarial prompts as ``too aggressive'' and roleplay as ``surface level.'' Once problem statements were locked, an autonomous paper pipeline received the same problem statement and ran without human intervention during the remaining session time. It serves only as a production-capacity limit case (\S\ref{sec:findings:autonomous}), not a controlled comparison.
\section{Findings}
\label{sec:findings}

Participants described faster generation without sufficient time for collective evaluation. Two recurring patterns appeared across all three independently operating teams; we interpret them as candidate mechanisms rather than identified causal effects.

\subsection{The consumption bottleneck}
\label{sec:findings:consumption}

\begin{figure}[t]
  \centering
  \includegraphics[width=0.5\linewidth]{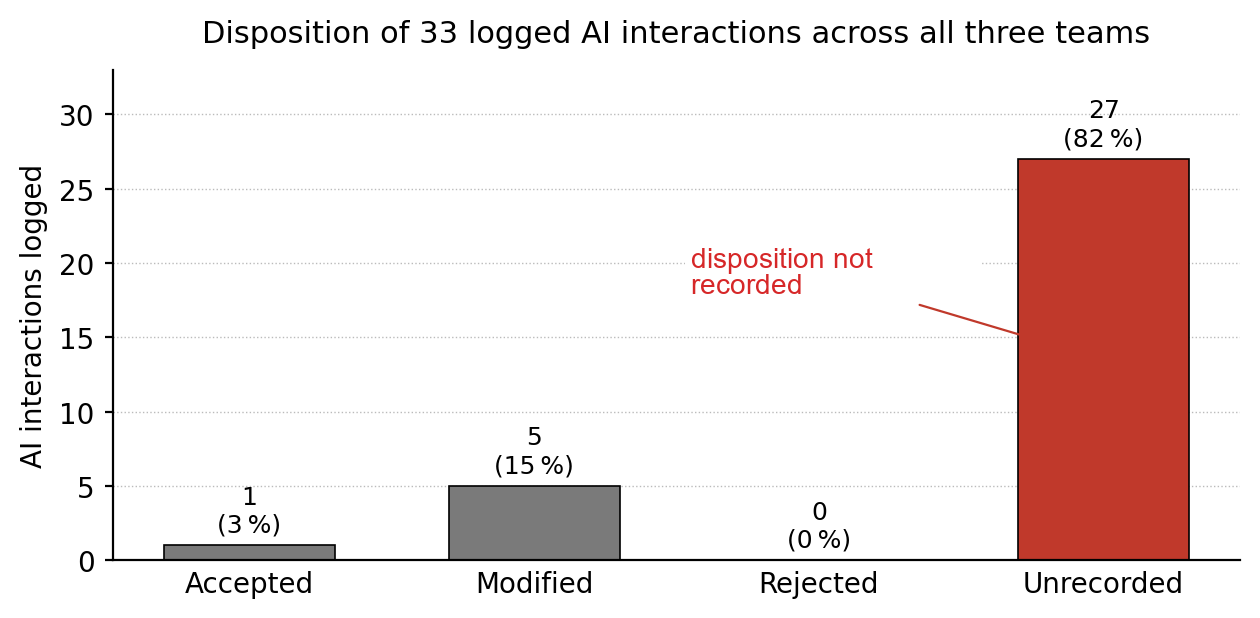}
  \caption{Disposition of 33 logged AI interactions: one (3\%) accepted, five (15\%) modified, none rejected, twenty-seven (82\%) unrecorded. Missing entries do not establish absent evaluation.}
  \label{fig:disposition}
\end{figure}

We define the consumption bottleneck as the gap between the rate at which AI produces outputs and the rate at which the team can evaluate them. ``Evaluate'' is the operative term: not read, but read closely enough to catch a substantively wrong claim. Under deadline pressure, this form of reading is the first activity to be deferred. The phenomenon was named by all three teams in survey or debrief, was followed by an observable behavioral correction in two of three teams (AI was reduced or unplugged at consequential moments), and registered in the instrument as a logging collapse: of 33 logged interactions, only one was accepted and twenty-seven received no disposition (Fig.~\ref{fig:disposition}). Two of the three teams independently produced the same verbatim self-diagnosis, ``plowed ahead instead of vetting,'' and reported tool-boundary friction in which copy/paste between Claude Code and the shared pipeline absorbed the time that would have been spent evaluating outputs.

\paragraph{What participants said.}
The adversarial debrief asked each team for the most friction-filled moment. Team~1: ``Content consumption and assimilation.'' Team~2: ``A lot of information to consume; combining outputs generated individually.'' Team~3, mid-session in a Stage~2 micro-survey: ``No time to consume AI generated results and make knowledgeable decisions on it.'' Three teams, three instruments, three nearly interchangeable formulations. These responses emphasize difficulty consuming outputs, rather than generation speed alone.

\paragraph{What teams did when they noticed.}
Two of three teams produced an observable behavioral correction. Team~1, approximately thirty minutes in, made an explicit decision: ``No AI tool used to design experiment.'' The team disabled the tools for the one stage in which incorrect outputs would be unrecoverable. Team~2 abandoned the adversarial prompt protocol mid-task and switched to a direct prompt because, in the participant's own debrief language, the adversarial style ``didn't work well'' under the available time. These adaptations suggest that participants responded to friction by reducing AI involvement or simplifying prompting; they do not establish that better prompts could not help. Both teams named the same compromise in the closeout debrief: ``due to time constraints, [we tried] to plow ahead instead of ensuring the quality of the work.''

\paragraph{Instrument behavior.}
Across all three teams, 33 interactions were logged: one accepted, five modified, twenty-seven with no disposition recorded (Fig.~\ref{fig:disposition}). Observer logging depended on a team-visible decision moment. Missing entries are consistent with private or deferred decisions, but may also reflect observer overload. They establish incomplete disposition records, not that every output was used without evaluation. Participant reflections and artifacts provide complementary evidence for the proposed bottleneck.

\paragraph{Mechanism.}
Using and evaluating generative AI imposes cognitive and metacognitive demands \citep{tankelevitch2024metacognitive,subramonyam2024gulf}; fluent text need not be grounded in meaning \citep{bender2021parrots}. Automation research documents attention-related complacency and demanding residual human roles \citep{parasuraman2010complacency,bainbridge1983ironies}. Explanations can increase acceptance regardless of correctness \citep{bansal2021whole}, but can reduce overreliance when they lower verification costs \citep{vasconcelos2023explanations}. Suggestion-verification time can be underestimated \citep{mozannar2024reading}. Our proposed extension is team-level: private contexts may make verification harder to distribute. Participants reported prioritizing progress over vetting. Whether this mechanism contributes to unevaluated research output remains untested.

\paragraph{Falsifiable form.}
Teams using private AI assistants under time pressure will evaluate a smaller share of AI outputs collectively than teams using shared-context assistants under equivalent pressure, or private assistants with more time.

\subsection{AI-induced research silos}
\label{sec:findings:silos}

The second failure is geometry, not volume. Unlike AI-mediated communication between people \citep{hancock2020aimediated}, private AI sessions need not expose their context to teammates. Under deadline, we hypothesize that attention may shift toward these sessions at the expense of shared workspace awareness \citep{gutwin2002workspace}. Private sessions can leave integration to individual judgment. Reduced diversity across AI-assisted stories \citep{doshi2024creativity} offers a possible analogy, not direct team-level evidence. Team~3 named it mid-session (``AI is multiplier for personal work not team work''); Team~1 theorized it in real time (``AI utility theorized to increase when working in silo''); Team~2 produced the most direct artifact, a Claude-drafted hypothesis summary ``with acronyms without any context, making it difficult to understand for other team members.'' Integration was performed, in Team~1's words, ``subjectively by a team member''; Team~3 logged idle time, ``waiting for the team picking the dataset.'' Asked what they would change, no team asked for stronger models or longer context. They asked to take the parallelism out.

\paragraph{Mechanism.}
The CSCW and team-cognition literatures offer interpretive foundations: grounding contributions in shared understanding \citep{clark1991grounding}, awareness of others' workspace activity \citep{gutwin2002workspace}, cognition distributed across people and artifacts \citep{hollan2000distributed}, and knowledge of who knows what \citep{wegner1987transactive}. Team-automation research motivates observability and directability \citep{klein2004ten,seeber2020machines}, while trust research emphasizes appropriate reliance \citep{lee2004trust}. We hypothesize that private Human-AI sessions can weaken these conditions when teammates cannot inspect the context; these sources do not establish that causal link.

\paragraph{Falsifiable form.}
Teams using AI under time pressure will show lower cross-person communication density and higher within-person AI interaction density than comparable teams without AI under equivalent pressure.

\subsection{Autonomous production: AI-AI limit case}
\label{sec:findings:autonomous}

The autonomous pipeline produced a complete draft on the same problem statement during the remaining session time, without human intervention. This demonstrates end-to-end production capacity, not superior research quality: neither its artifact nor the team artifacts received independent quality scores. The run is a mechanism probe, not a control condition or a statistical comparison. It illustrates the generation--evaluation asymmetry with in-session human evaluation removed, rather than its cost reduced. Verification is deferred to downstream readers unless an explicit review process is provided. Autonomous execution combined with inspectable team review remains compatible with Human-AI-Human (\S\ref{sec:alternatives}); this run did not test that configuration.

\subsection{Convergence}
\label{sec:findings:convergence}

\begin{figure}[t]
  \centering
  \includegraphics[width=0.85\linewidth]{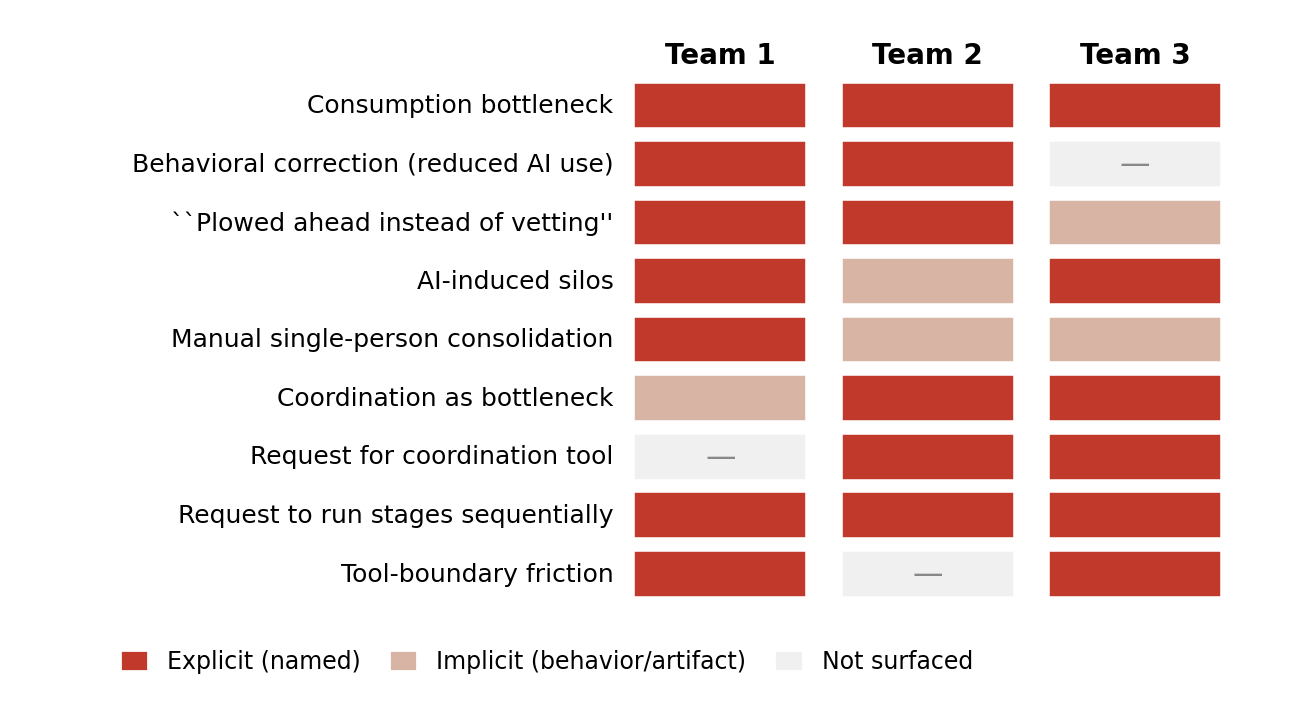}
  \caption{Convergence across three independent teams. \emph{Explicit} = a participant or observer named the failure directly; \emph{implicit} = it surfaced as behavior or artifact rather than as a stated claim.}
  \label{fig:convergence}
\end{figure}

Three independent teams, same clock, same tools, no shared communication channel, surfaced the same failures (Fig.~\ref{fig:convergence}). We treat this as within-session convergence, not independent experimental replication or statistical generalization. Shared organizational context and session conditions limit the inference.

\paragraph{From the closed room to the enterprise floor.}
Participants described the session as resembling their normal week ``with the time budget compressed.'' This motivates, but does not establish, relevance to enterprise research. Time pressure, role clarity, workflow maturity, and missing shared infrastructure may jointly explain the patterns. Faster generation could increase evaluation demands, while more reliable models or practiced workflows could reduce them. Controlled field studies must distinguish these possibilities. Section~\ref{sec:gap} turns the observations into a testable design agenda.
\section{The Human-AI-Human design category}
\label{sec:gap}

Participants named the missing tool before we did: ``a better coordination tool so that context from different team members can be used to make better decisions from Claude''; ``a way to share sessions to have a final consolidation''; two teams asked for Claude Code to be ``integrated to the tool we provided'' to remove copy/paste; every team asked to take parallelism out and run stages sequentially.

\subsection{The category, defined}

Here, \textbf{Human-AI} denotes individual-centered assistance \citep{vaithilingam2022expectation,lee2022coauthor,heer2019agency}. Human-first drafting, AI-first proposals, and repeated refinement all fit this topology. \textbf{AI-AI} denotes autonomous runs without in-loop human evaluation \citep{lu2024aiscientist,yamada2025aiscientistv2}; it defers verification rather than eliminating its cost. This labels an execution mode, not every use of a framework: AutoGen and Agent Laboratory also support human input \citep{wu2023autogen,schmidgall2025agentlab}.

\textbf{Human-AI-Human} makes the team the unit of interaction and decision-making. AI selectively connects researchers' working contexts, preserves provenance, exposes conflicts, and routes claims for collective evaluation before consequential decisions. This is an interaction topology, not a fixed sequence: either humans or AI may initiate, and interaction may be single-pass or iterative. Unlike sharing only a final document, the category makes assumptions, intermediate evidence, and acceptance decisions inspectable across people. Existing collaboration platforms can implement it through shared memory, expertise-aware routing, and team-visible decision gates. Key challenges include permission-respecting context sharing, multi-user reasoning, and avoiding information overload.

\subsection{Design requirements as a falsifiable agenda}
\label{sec:gap:requirements}

Table~\ref{tab:requirements} derives seven modular hypotheses from observed gaps (\S\ref{sec:findings}), not validated prescriptions. They can span assistants, workspaces, and workflows; no single tool must implement all seven. Their benefits and burdens require testing separately and in combination.

\begin{table}[t]
\centering
\small
\caption{The Human-AI-Human design agenda. Each requirement maps to specific evidence in \S\ref{sec:findings} and a prediction a future study can refute.}
\label{tab:requirements}
\begin{tabular}{@{}p{0.5cm} p{3.6cm} p{4.6cm} p{5.6cm}@{}}
    \toprule
    \textbf{ID} & \textbf{Requirement} & \textbf{Failure addressed} & \textbf{Falsifiable prediction} \\
    \midrule
    R1 & Shared context substrate across sessions and tools & Idle Claude Code session; cross-tool copy/paste & Tools sharing a substrate show higher cross-person reuse of AI outputs \\
    R2 & Conflict and redundancy detection across workstreams & Parallel-session fragmentation in all teams & Tools that flag converging or contradictory sessions reduce late integration failures \\
    R3 & Forced shared evaluation checkpoints & 27/33 dispositions unrecorded (Fig.~\ref{fig:disposition}) & Team-level forcing functions raise the share of AI outputs passing team-visible evaluation \\
    R4 & Disposition logging team-visible by default & 82\% blank rate on external observer log under load & Built-in logging produces a more complete record than external annotation \\
    R5 & Role-aware routing of AI output to relevant expertise & Team~2 acronym-laden hypothesis summary & Routing to expertise-matched teammates with a translation pass reduces incomprehensibility reports \\
    R6 & Decision-ownership tracking & First 10 min: research question silently delegated to Claude & Tools recording decision ownership produce more contested mid-session course corrections \\
    R7 & Epistemic load management & Consumption bottleneck across all teams & Throttling unreviewed AI output (batching, ack gates) reduces unevaluated content in the artifact \\
    \bottomrule
\end{tabular}
\end{table}

The minimum initial intervention combines shared cross-person context (R1) and collective evaluation checkpoints (R3). Compare it against existing tools with matched facilitation, measuring cross-person reuse, reviewed-output share, integration failures, and unreviewed content. This tests the combination, not each component's effect.

\subsection{Alternative positions and boundary cases}
\label{sec:alternatives}

\paragraph{Existing platforms may suffice.} AutoGen \citep{wu2023autogen} and MetaGPT \citep{hong2024metagpt} provide agent coordination. Combined with shared editing, such frameworks could support Human-AI-Human if they connect working contexts and collective decisions. The distinction is functional, not a claim that a separate product is necessary.

\paragraph{Why not autonomous execution and shared review?} Automated discovery and experimentation \citep{romera2024funsearch,boiko2023coscientist} can feed one shared synthesis, reducing private silos and immediate attention costs. A summary alone may hide consequential assumptions. This becomes Human-AI-Human when provenance, uncertainty, alternatives, and evidence remain inspectable and the team can challenge or revise decisions at consequential checkpoints. Autonomous execution is compatible with collective agency.

\paragraph{Better workflows may suffice.} Clearer roles, facilitation, practice, improved prompts, or more time could mitigate the observed problems; our pilot does not exclude this explanation. Follow-up studies should compare these improvements with shared-context AI, rather than assume tooling is the sole cause. Individual productivity findings \citep{peng2023copilot} do not settle which approach best supports collective evaluation.
\section{Limitations and what would change our minds}
\label{sec:limitations}

This single-organization, opportunistic pilot involved twelve task-performing researchers in a seventeen-person, 100-minute session. It generates hypotheses, not causal, prevalence, or effect-size estimates. Without a no-AI control, it cannot separate AI-specific amplification from pressure, role clarity, or workflow maturity. Neither team artifacts nor the single autonomous draft received independent quality scores. Missing dispositions may reflect both private decisions and logging limitations. The tool snapshot is time-specific; fundamental research, longer projects, and non-research tasks remain untested settings.

Longitudinal, multi-organization studies should vary time budgets, workflow training, repeated sessions, and individual/shared/no-AI conditions, recording baseline AI proficiency. Outcomes should include communication, integration and verification time, collective review, and independently assessed research quality. Recovery through practice or facilitation alone would weaken the tooling explanation; no added benefit from shared-context systems would challenge the proposed intervention. The present diagnostic study neither implemented nor validated the agenda in \S\ref{sec:gap:requirements}.

\section{Conclusion}
\label{sec:conclusion}

Our pilot surfaced recurring consumption and integration difficulties when researchers used personal AI tools under deadline pressure. It does not show that AI caused every difficulty or reduced research quality. It motivates a shift in design and evaluation: from individual throughput alone toward inspectable collective reasoning. Human-AI-Human names this team-centered topology, compatible with existing platforms and autonomous execution. Shared context and collective evaluation checkpoints offer a concrete starting point for testing whether better team processes translate into better research.

{\small
    \bibliographystyle{plainnat}
    \bibliography{reference}
}

\end{document}